\documentclass[runningheads]{llncs}
\usepackage[T1]{fontenc}
\usepackage{graphicx}
\usepackage{subcaption}
\usepackage{wrapfig}

\usepackage{amssymb}
\usepackage[none]{hyphenat}
\usepackage{hyperref}      
\usepackage{color}

\usepackage{amsmath}
\usepackage{mathtools}  
\usepackage{stackrel}   
\usepackage{booktabs}

\usepackage{listings}
\usepackage{xcolor}
\definecolor{codegreen}{rgb}{0,0.6,0}
\definecolor{codegray}{rgb}{0.5,0.5,0.5}
\definecolor{codepurple}{rgb}{0.58,0,0.82}
\definecolor{backcolour}{rgb}{0.95,0.95,0.92}

\lstdefinestyle{yamlstyle}{
    backgroundcolor=\color{backcolour},
    commentstyle=\color{codegreen},
    keywordstyle=\color{magenta},
    numberstyle=\tiny\color{codegray},
    stringstyle=\color{codepurple},
    basicstyle=\ttfamily\scriptsize,    
    breakatwhitespace=false,
    breaklines=true,
    captionpos=b,
    keepspaces=true,
    numbers=left,
    numbersep=5pt,
    showspaces=false,
    showstringspaces=false,
    showtabs=false,
    tabsize=2,
}

\definecolor{backcolour}{rgb}{0.95,0.95,0.92}
\definecolor{codegreen}{rgb}{0,0.6,0}
\definecolor{magenta}{rgb}{1,0,1}
\definecolor{codegray}{rgb}{0.5,0.5,0.5}
\definecolor{codepurple}{rgb}{0.58,0,0.82}

\lstdefinestyle{prismstyle}{
    backgroundcolor=\color{backcolour},
    commentstyle=\color{gray}\itshape,
    keywordstyle=\color{blue}\bfseries,
    numberstyle=\tiny\color{orange},
    stringstyle=\color{purple},
    identifierstyle=\color{black},
    basicstyle=\fontfamily{cmtt}\footnotesize,
    breakatwhitespace=false,
    breaklines=true,
    captionpos=b,
    keepspaces=true,
    numbers=left,
    numbersep=5pt,
    showspaces=false,
    showstringspaces=false,
    showtabs=false,
    tabsize=2,
    morekeywords={dtmc, module, endmodule, const, formula, rewards, endrewards},
    morecomment=[l]{//}, 
}

\begin{document}
\title{Probabilistic Modeling of Spiking Neural Networks with Contract-Based Verification}
\titlerunning{Probabilistic Modeling and Verification of Spiking Neural Networks}
%
\author{Zhen Yao\inst{1}\orcidID{0009-0007-4848-7689} \and
Elisabetta De Maria\inst{1}\orcidID{0000-0001-7116-9629} \and
Robert De Simone\inst{2}\orcidID{0000-0002-3123-7591}}
\authorrunning{Z. Yao et al.}
%
\institute{Université Côte d'Azur, I3S, CNRS, France\\
\email{zhen.yao@etu.univ-cotedazur.fr, edemaria@i3s.unice.fr}\and
Centre Inria d'Université Côte d'Azur, France\\
\email{robert.de\_simone@inria.fr}}

\maketitle              
\begin{abstract}
Spiking Neural Networks (SNN) are models for "realistic" neuronal computation, which makes them somehow different in scope from "ordinary" deep-learning models widely used in AI platforms nowadays. SNNs focus on timed latency (and possibly probability) of neuronal reactive activation/response, more than numerical computation of filters.
So, an SNN model must provide modeling constructs for elementary neural bundles and then for synaptic connections to assemble them into compound data flow network patterns. These elements are to be parametric patterns, with latency and probability values instantiated on particular instances (while supposedly constant "at runtime"). Designers could also use different values to represent "tired" neurons, or ones impaired by external drugs, for instance.
One important challenge in such modeling is to study how compound models could meet global reaction requirements (in stochastic timing challenges), provided similar provisions on individual neural bundles. A temporal language of logic to express such assume/guarantee contracts is thus needed. This may lead to formal verification on medium-sized models and testing observations on large ones.
In the current article, we make preliminary progress at providing a simple model framework to express both elementary SNN neural bundles and their connecting constructs, which translates readily into both a model-checker and a simulator (both already existing and robust) to conduct experiments.

\end{abstract}
\section{Introduction}

Spiking Neural Networks constitute the third generation of neural network models, marrying computational efficiency with biological plausibility. Unlike traditional rate‐based Artificial Neural Networks (ANNs), SNNs employ event‐driven spikes—discrete voltage pulses—to encode and transmit information, focusing on the timed latency and probabilistic nature of neuronal activation. This mechanism greatly reduces energy consumption by activating neurons only when spikes occur, and naturally captures temporal dynamics, making SNNs well-suited for tasks involving time-varying inputs such as signal processing, motor control, and event-based vision~\cite{maass1997networks}.

Early demonstrations have shown that SNNs can match ANNs on both classical benchmarks (e.g., MNIST, N-MNIST, N-Caltech101)~\cite{kabilan2021neuromorphic,LeCun1998} and more challenging datasets like ImageNet and CIFAR-10~\cite{Sengupta2019}, while preserving their energy-saving advantages. Beyond performance, the bio-inspired nature of SNNs renders them invaluable for neuroscientific inquiry, offering a computational window into the dynamics of real neural circuits. Central to SNN performance is the choice of neuron model, which must balance biological realism against computational tractability. Leaky Integrate-and-Fire (LI\&F) neurons afford efficiency but neglect some dynamics; biophysically detailed Hodgkin-Huxley models capture a wealth of ionic processes at the cost of prohibitive computational load~\cite{hodgkin1952quantitative}. Recent advances—in particular, Generalized Leaky Integrate-and-Fire (GLIF)~\cite{Teeter2018} and Noisy LI\&F~\cite{Maio2004} models—have introduced stochastic thresholds and membrane-potential noise to better mimic cortical variability. However, existing variants still fall short of emulating absolute and relative refractory behaviors observed in biological neurons.

In this paper, we introduce the \textbf{Refractory-evolve Probabilistic Leaky Integrate-and-Fire (RP-LI\&F)} neuron model based on previous Boolean Probabilistic Spiking Neural Network~\cite{DeMaria2018}. RP-LI\&F embeds discrete-time refractory dynamics and probabilistic spike generation into the LI\&F framework, yielding a compact yet biologically grounded unit. We support RP-LI\&F with a domain-specific format, Spiking Neural Networks Representation File (SNN-RF), that automatically produces discrete-time Markov chain models for formal verification in PRISM and high-performance simulations in Nengo~\cite{Bekolay2014}.

SNN models require parametric constructs for elementary neural bundles and synaptic connections, enabling the assembly of complex data-flow network patterns. These constructs must support the instantiation of latency and probability parameters, which remain constant during runtime but can vary to simulate conditions such as neuronal fatigue or pharmacological effects. A key challenge is ensuring that compound SNN models meet global reaction requirements under stochastic timing constraints, given similar provisions for individual neural bundles. This necessitates a temporal logic language to express assume/guarantee contracts, facilitating formal verification for medium-sized models and testing through simulation for larger ones.

 This preliminary work aims to develop a brain-like neural network by archetypes~\cite{DeMaria2022}, which are specific neuronal micro-circuits, that can be combined in the future to form complicated, brain-inspired systems, distinct from second-generation networks focused solely on data processing efficiency. The remainder of this paper is structured as follows. Section~\ref{sec:neuron} introduces our RP-LI\&F neuron model. Section~\ref{sec:tools} describes the PRISM model checker with the logic PCTL and the Nengo simulator. Section~\ref{sec:setup} presents how we automatically generate a network model in both Nengo and PRISM via SNN-RF. Section~\ref{sec:cases} details case studies demonstrating RP-LI\&F in contralateral inhibition and convergent excitatory circuit, two fundamental archetypes~\cite{DeMaria2022}. Finally, Section~\ref{sec:conclusion} concludes and outlines directions for future work. The code can be found at \url{https://github.com/zxy1553/RP-LIF.git}.

\section{Refractory-Evolve Probabilistic LI\&F Model}\label{sec:neuron}
Real neurons maintain a resting voltage that slowly drifts back after synaptic inputs, fire when they reach a threshold, exhibit random channel‐driven fluctuations, and pause briefly after each spike before regaining full excitability.  In our RP‐LI\&F abstraction, the leak factor \(r=\exp(-\Delta t/\tau_m)\) models the passive return to rest, the threshold \(\tau\) marks the voltage for spike initiation, the membrane potential \(p(t)\) sums incoming spikes with exponential decay, discrete probability levels \(\{p_i\}\) capture channel noise that makes firing stochastic, the absolute‐refractory countdown \(\mathrm{ARP}\) enforces a short window of zero spiking, and the relative‐refractory countdown \(\mathrm{RRP}\) with scaling factor \(\alpha\) represents the gradual restoration of firing probability after an action potential.

We define the \textbf{Refractory-evolve Probabilistic Leaky Integrate-and-Fire} (RP-LI\&F) neuron as a tuple \(v = (\tau, r, p, y, s, \texttt{aref}, \texttt{rref})\), where:
\begin{itemize}
  \item \(\tau \in \mathbb{N}\): reference firing threshold
  \item \(r \in [0,1]\cap\mathbb{Q}\): leak factor, governing exponential decay of potential.
  \item \(p\colon \mathbb{N}\to\mathbb{Q}_{0}^{+}\): membrane potential at discrete time \(t\), with
    \begin{equation}\label{eq:membrane_potential}
        p(t) = 
        \begin{cases}
            \displaystyle\sum_{i=1}^{m} w_i\,x_i(t), 
            & \text{if } y(t-1)=1,\\[1ex]
            \displaystyle\sum_{i=1}^{m} w_i\,x_i(t)\;+\;r\,p(t-1), 
            & \text{otherwise},
        \end{cases}
    \end{equation}
    where \(p(0)=0\), \(m\) is the number of inputs, \(w_i\) the synaptic weight, and \( x_i(t) \in \{0,1\} \) the incoming spike.
    \item \(y\colon \mathbb{N}\to\{0,1\}\): output spike function, \(y(t)=1\) iff a spike is emitted at \(t\).
    \item \(s\in\{0,1,2\}\): state flag, taking values \(0\) for Normal period, \(1\) for Absolute refractory period, and \(2\) for Relative refractory period.
    \item \(\texttt{aref}\in\{0,\dots,\texttt{ARP}\}\): absolute-refractory countdown,  
    \item \(\texttt{rref}\in\{0,\dots,\texttt{RRP}\}\): relative-refractory countdown.
\end{itemize}

\subsection{Spike-Probability Law}

Let \(\Delta(t)=p(t)-\tau\). Let us discretize \(\Delta\) into \(2k+1\) intervals \(\{-l_k,\dots,0,\dots,l_k\}\). We define a base firing probability as follows:
\[
P_{\mathrm{base}}(y(t)=1) \;=\;
\begin{cases}
1, & \Delta(t)\ge l_k,\\
p_{2k}, & l_{k-1}\le\Delta(t)<l_k,\\
\vdots\\
p_{k+1}, & 0\le\Delta(t)<l_1,\\
p_k, & -l_1\le\Delta(t)<0,\\
\vdots\\
p_1, & -l_k\le\Delta(t)<-l_{k-1},\\
0, & \Delta(t)<-l_k.
\end{cases}
\]
Here, \(\{p_i\}\) are predefined probabilities over \(\Delta/\tau\). Then the actual spike probability is
\[
P\bigl(y(t)=1\bigr)
=\begin{cases}
0, & s=1,\\
\alpha\;P_{\mathrm{base}}(y(t)=1), & s=2,\\
P_{\mathrm{base}}(y(t)=1), & s=0,
\end{cases}
\]
with \(\alpha\in[0,1]\) shrinking relative‐refractory firing.

\subsection{Refractory Dynamics}
\paragraph{Absolute Refractory Period (\(s=1\))}  
For this step after a spike, \(y(t)=0\) always, \(p(t)\equiv0\), and \(\texttt{aref}\!\to\!\texttt{aref}-1\) each step.  When \(\texttt{aref}=0\), transition to \(s=2\).

\paragraph{Relative Refractory Period (\(s=2\))}  
For this step, the neuron may fire at a reduced rate \(\alpha\, P_{\mathrm{base}}\).  \(\texttt{rref}\) decrements until zero; if a spike occurs, reset \(\texttt{aref}\) and re-enter \(s=1\), else when \(\texttt{rref}=0\), return to \(s=0\).

\begin{figure}[h]
  \centering
  \includegraphics[width=0.75\linewidth]{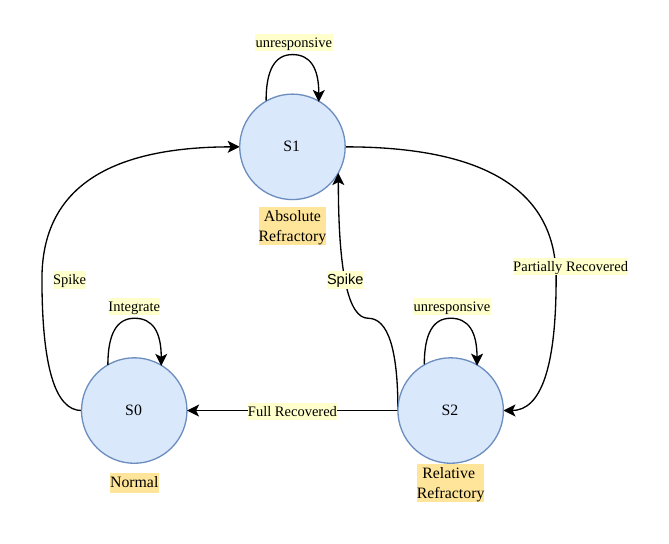}
  \caption{RP-LI\&F neuron state transitions.}
  \label{fig:automata}
\end{figure}

Figure~\ref{fig:automata} illustrates the three-state finite-state machine governing transitions among Normal, Absolute, and Relative refractory periods based on spike events and counters.

\section{SNN Toolchain Overview}\label{sec:tools}
This section introduces the PRISM model checker for formal verification and Nengo for simulation. PRISM allows one to model systems with probability aspects and to verify dynamic properties. Nengo leverages the Neural Engineering Framework (NEF) to build and simulate mechanistic neuronal circuits. Together, they offer a comprehensive approach to designing, validating, and analyzing SNNs.

\subsection{PRISM}
PRISM~\cite{Kwiatkowska2011} is a powerful framework designed for probabilistic modeling and verification, enabling the analysis of diverse probabilistic systems such as discrete and continuous-time Markov chains (DTMCs and CTMCs), Markov Decision Processes (MDPs). In addition to modeling, PRISM supports the formal specification of system behavior using probabilistic temporal logics, including PCTL~\cite{Hansson1994} (Probabilistic Computation Tree Logic).

\subsubsection{Syntax and modeling}
We model our SNN as a DTMC by representing each neuron or population as a PRISM module with local variables capturing membrane potential and spike events. Guarded commands of the form:
\[
[\textit{action}] \, \textit{guard} \to \textit{prob}_1 : \textit{update}_1 + \cdots + \textit{prob}_n : \textit{update}_n;
\]
encode probabilistic spiking and rest behaviors. Synchronization labels \textit{action} ensure coordinated updates across neurons at each time step. Reward structures are added to track metrics such as total spikes or energy consumption, enabling queries of the form:
\(\mathbf{R}\{"y"\}=? \, [\mathbf{C}^{\leq 100}]\)
to compute expected spike counts within a time bound.

\subsubsection{PCTL and Model Checking}

PRISM employs PCTL to define the DTMCs with probabilistic and temporal constraints. PCTL extends CTL with a probabilistic path quantifier (e.g. \(P_{\geq p}[F\,\phi]\) and time bounds.


All quantifiers except $\mathbf{X}$ support time-bounded variants (e.g., $\mathbf{F}^{\leq t} \, \phi$). The $\mathbf{P}=? \, [\textit{prop}]$ form computes the exact probability of a property. 

In our workflow,  we automatically generate the PRISM model from a high-level description of the network topology: connectivity matrices, synaptic weights, and firing thresholds are parsed to instantiate modules and commands. We then specify PCTL queries for key properties and invoke PRISM's model checker to compute exact probabilities.

\subsection{Nengo}
Nengo is a neural modeling and simulation platform based on the NEF, which uses populations of spiking neurons to represent and transform vectors mechanically~\cite{DeWolf2020}. Users define networks via a Python API, specifying Ensembles (populations), Connections (transformations), and Probes (recordings).

We construct each SNN circuit in Nengo to align with the PRISM model: ensembles correspond to PRISM modules, and synaptic filters model transition probabilities. Simulation run on CPU or Liohi hardware generates spike trains and state trajectories over time. Probe data is exported to show how neurons in the neural network work. Both the DTMC abstraction and the NEF implementation are important to ensure our neural network works as we suppose.

\section{Automated SNN generation and modeling}\label{sec:setup}
In this section, we present a structured approach to modeling SNNs, encompassed in a dedicated representation file format (SNN-RF), itself compliant for translation to both PRISM and Nengo
analysis frameworks.  By defining both individual neuron archetypes and combining network properties in YAML-based representation files, we enable the automatic generation of models for both simulation and formal verification. This method simplifies the configuration of complex neural architectures, such as those based on our custom Refractory-evolve Probabilistic Leaky Integrate-and-Fire (RP-LI\&F) neurons, while ensuring consistency across different modeling tools.

We begin by outlining the structure and purpose of these representation files, followed by detailing how they are used to generate PRISM and Nengo models. Finally, we provide insights into the implementation details within each tool, highlighting the custom neuron models and their behaviors.

\subsection{Representation Files}
The SNN-RF format essentially provides a parametric description of the elementary neuron archetypes (in the present case RP-LI\&F), and the synaptic connections to combine and associate them in larger structures.
Distinct neurons can be instantiated by fixing local values for the parameters (timing and probability thresholds of reactive spikes). The simple YAML form of SNN-RF is designed to allow simple and correct
translation into PRISM and Nengo, and also to promote easily readable configuration of parameters and network topologies through lists and dictionaries.

Importantly, the single source ensures that the resulting PRISM and Nengo descriptions have identical behaviors and therefore allows cross-validation between simulation and model-checking results.
Currently, correctness requirements are expressed only directly using PCTL probabilistic temporal logic in PRISM, but in the future, our SNN-RF format should also provide simple parametric requirement patterns to be translated both in PCTL for PRISM and as runtime observers for Nengo.

Below is a snippet from our network representation file:

\begin{lstlisting}[style=yamlstyle,caption=Spiking Neuron Network Representation File, label=lst:yaml_config_structure]
network:
    name: "STRING"  # type of neural network
    simulate:
        steps: INTEGER
    inputs:
        - id: INTEGER
          value: INTEGER
          ......
    n_neurons:
        - id: INTEGER
          # More properties for neurons in the network
          ......
    edges:
        - from:
            type: STRING
            id: INTEGER
          to:
            type: STRING
            id: INTEGER
          weight: INTEGER
        ......
\end{lstlisting}

\subsection{Neuron for PRISM}
Our SNN comprises multiple modules, collectively forming a complete DTMC model. Typically, the model includes an input layer, a synapse layer, and an output layer. Each neuron receives a single input signal, and all inputs are grouped into the Input module. Neurons are assigned distinct weights for their binary inputs (restricted to 0 or 1). Neuron modules are named by combining their type with a unique identifier (e.g., Neuron1), enabling the expansion of our model into multi-layer architectures. Neurons across different layers are interconnected through synapses implemented as transfer modules. The Input, neuron, and transfer modules serve as the foundational building blocks of our network, enabling the generation of various archetypes. The core neuron behavior is encapsulated in a Neuron module:
\begin{lstlisting}[style=prismstyle,breaklines=true, linewidth=0.9\textwidth]
module Neuron1
  aref : [0..ARP] init 0; //absolute refractory
  rref : [0..RRP] init 0; //relative refractory

  s: [0..2] init 0;  // state
  y: [0..1] init 0;  // spike
  p: [-500..500] init P_rest;  // membrane potential

  [to] s = 0 & y = 0 & p < threshold1 -> 
       1.0: (y' = 0) & (p' = newPotential);
  [to] ...
  [to] s = 0 & y = 0 &
       p >= threshold5 & p < threshold6 -> 
       0.5: (y' = 0) & (p' = newPotential) + 
       0.5: (y' = 1);
  [to] ...
  [to] s = 0 & y = 0 & p > threshold10 -> 
       1.0: (y' = 1);
endmodule
\end{lstlisting}

The neuron spikes (\(y=1\)) when its potential exceeds a threshold, resetting to a basic value and entering an absolute refractory period (\(s=1\)). We implement the update in Equation~\ref{eq:membrane_potential} by first taking the weighted sum plus the leak term, rounding it down to an integer, and then clipping the result to lie between \texttt{MAX} and \texttt{MIN}, where \texttt{MAX} and \texttt{MIN} denote the upper and lower bounds of the discretized membrane potential:
\(\text{newPotential}
  = \max\Bigl(\min\bigl(\lfloor \sum_{i=1}^m w_i x_i + r \cdot p\bigr\rfloor,\;\texttt{MAX} \Bigr),\;\texttt{MIN}\Bigr),
\)
 which ensures our discrete membrane potential remains within the specified bounds. Moreover, a rewards structure counts spikes:
\begin{lstlisting}
rewards "spike1_count"
  y1 = 1 : 1;
endrewards
\end{lstlisting}
enabling precise analysis of network activity.

\subsection{Neuron for Nengo}

Nengo’s standard neurons lack probabilistic spiking and refractory periods, so we introduce ProbNeuron, based on our RP-LI\&F model, to address these limitations. Using Nengo’s API, we build networks with single-neuron ensembles (encoder=1) and a step delay to align with PRISM models. We probe spike activity, membrane potential, and state, synchronizing the time step \emph{dt} with PRISM’s Markov chain for consistent, high-precision verification and simulation.

The behavior of our RP-L\&IF neuron is encapsulated in the ProbNeuron class:

\begin{lstlisting}[language=Python]
class ProbNeuron(nengo.neurons.NeuronType):
    def __init__(self, P_rth=10, P_rest=0,
     r=0.7, ARP=2, RRP=5, alpha=0.08, dt=0.001):
        super().__init__()
        self.P_rth = P_rth  # threshold
        self.ARP = ARP      # absolute refractory
        self.RRP = RRP      # relative refractory
        self.alpha = alpha  # shrinking factor
        self.dt = dt        # time step

    def step(self, dt, J, output, **state):
        s = state["s"]  # state
        p = state["p"]  # membrane potential
        y = state["y"]  # spike
        if s == 0:  # normal state
            p = np.floor(J + self.r * p)
            prob = self.get_spike_prob(p, is_rrp=False)
            y = 1 if np.random.rand() < prob else 0
        # ... (handle refractory period)
        output[...] = y
\end{lstlisting}

\section{Case Study}\label{sec:cases}

We validate our framework with one RP-LI\&F neuron, two archetypes: contralateral inhibition and convergent excitatory circuit. These demonstrate our approach for probabilistic verification and simulation, using SNN-RF to generate consistent DTMCs and Nengo Ensembles with shared \textit{dt} time steps, ensuring precise cross-domain validation (Figures~\ref{fig:ci}, \ref{fig:cec}). All experiments were performed on a machine with an \texttt{Intel(R) Core(TM) i5-8300H CPU @ 2.30\,GHz}, 16\,GB RAM, running \textsc{Arch Linux} (kernel version \texttt{6.14.6.arch1-1}).



\subsection{RP-L\&IF Neuron}
\begin{figure}[ht]
    \centering
    \begin{minipage}{0.45\textwidth}
        \includegraphics[width=\linewidth]{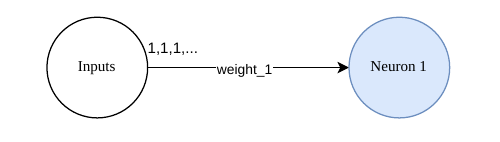}
        \subcaption{RP-LI\&F Neuron}
        \label{fig:one}
    \end{minipage}
    \hfill
    \begin{minipage}{0.45\textwidth}
        \includegraphics[width=\linewidth]{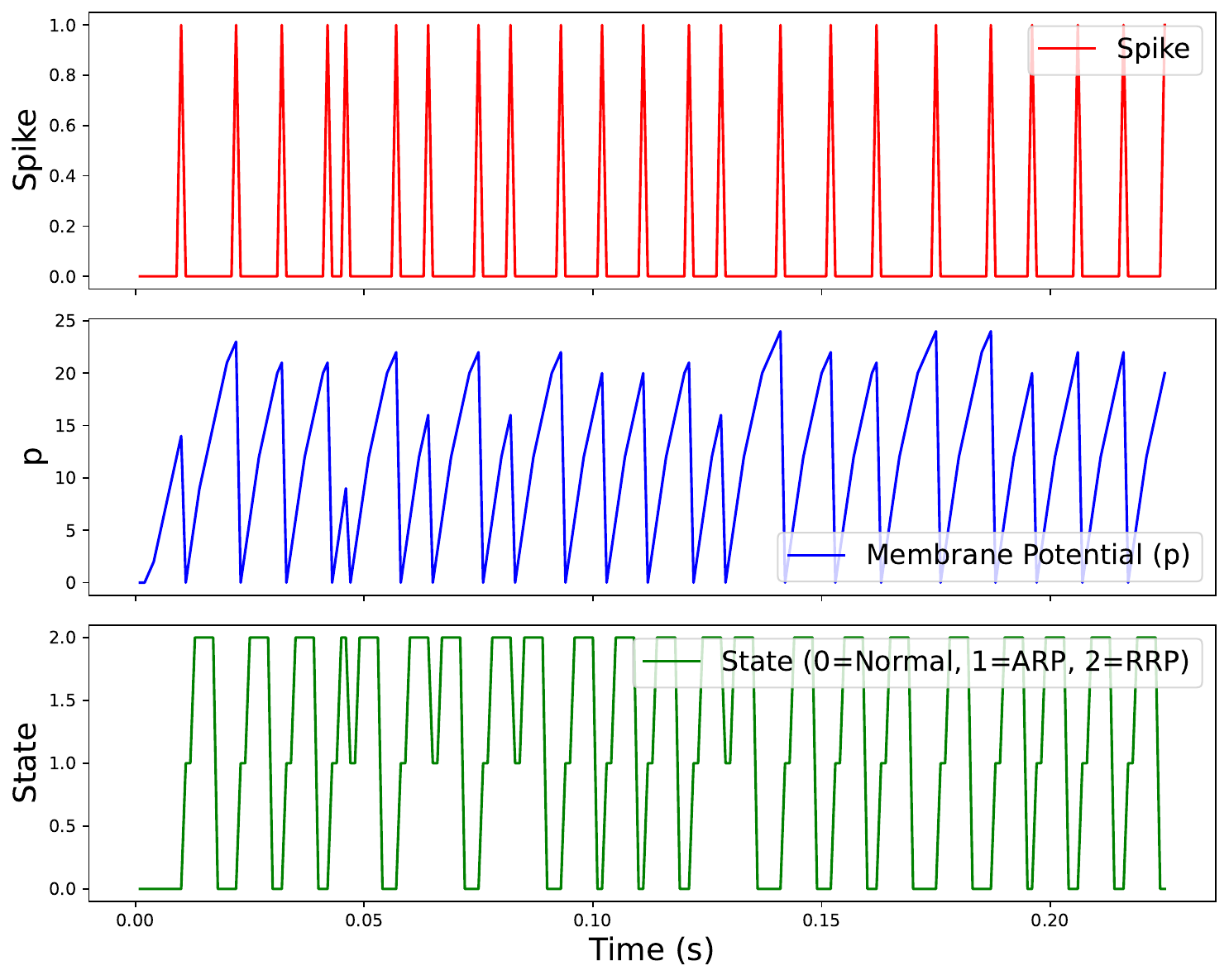}
        \subcaption{RP-LI\&F Neuron}
        \label{fig:on}
    \end{minipage}
    \caption{RP-LI\&F Neuron and Simulation (the X-axis represents time, and the Y-axis, from top to bottom, represents spike, membrane potential, and states)}
\end{figure}

Before we go into those two cases, first, we present the verification and simulation of only one RP-LI\&F neuron to understand better how our neuron works. Here, we checked the correct implementation of the RP-LI\&F neuron. The input of this neuron is a constant one (see Figure~\ref{fig:one}). 

\subsubsection{PCTL properties and results}
We validated the fundamental characteristics of the RP-LI\&F neuron, specifically its transition rules and spiking probabilities, which are essential for all subsequent developments.

\(P_{1}:\,P_{>=1} [ G\,((y_{1}=1) \longrightarrow (X\,(s=1)))]\) verified that whenever a spike occurs, the very next state is the absolute refractory period, in which the neuron does not emit.

\(P_{2}:\,P_{>=1} [ G\,((s=1 \wedge \texttt{aref=0}) \longrightarrow (X\,(s=2 \wedge \texttt{rref=RRP}) )  ) ] \) asserts that when \texttt{aref} reaches 0 in state 1, the next state is 2 with \texttt{rref} reset to its maximum.

\(P_{3}:\,P_{=?} [ ((p > threshold5) \wedge (p <= threshold6)) \longrightarrow (X\, (y=1)) ] \) checks the correctness of one of our probabilistic transitions. All the others can also be proved to be 1 in probability.

Table~\ref{tab:results-ci-props} shows the outcomes and execution time of Properties \(P_{1}\) to \(P_{3}\). 

\begin{table}[htbp]
\centering
\caption{Verification Results of RP-LI\&F Neuron}
\label{tab:results-ci-props}
\resizebox{\textwidth}{!}{%
\begin{tabular}{@{}cccc@{}}
\toprule
 \textbf{Properties} & \textbf{Specifications} & \textbf{Outcomes} & \textbf{Time}\\
\midrule
\(P_1\)
& $P_{>=1} [ G\,((y_{1}=1) \longrightarrow (X\,(s=1)))]$
& True
& 45 ms\\
\midrule
\(P_2\)
& $P_{>=1} [ G\,((s=1 \wedge aref=0) \longrightarrow (X\,(s=2 \wedge rref=RRS) )  ) ]$
& True
& 19 ms\\
\midrule
\(P_3\)
& $P_{=?} [ ((p > threshold5) \wedge (p <= threshold6)) \longrightarrow (X\, (y=1))] $
& 0.5
& 9 ms\\
\bottomrule
\end{tabular}%
}
\end{table}

\subsubsection{Nengo Simulation}
The Nengo simulation results (see Figure~\ref{fig:on}) illustrate a single neuron’s spiking behavior and state transitions.

\subsection{Contralateral Inhibition}
\begin{figure}
    \centering
    \includegraphics[width=0.75\linewidth]{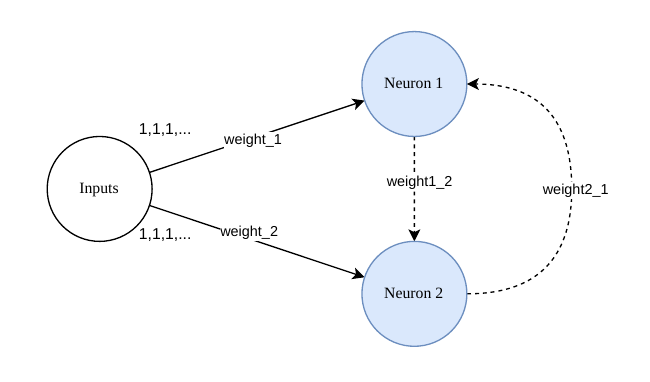}
    \caption{Contralateral Inhibition}
    \label{fig:ci}
\end{figure}

Contralateral Inhibition models neural populations that mutually suppress activity, a mechanism critical to sensory processing. They provide an elementary computation that can be used for decision-making (not only simple forms of behavioral choices but also more complex choices)~\cite{Koyama2018}. As another example, active and passive fear responses are mediated by distinct and mutually inhibitory central amygdala neurons~\cite{Fadok2017}. Here we provide a network that consists of one input always being 1, and two RP-L\&IF neurons which inhibit each other, expecting behavior in winner takes all, see Figure~\ref{fig:ci}, where the solid line represents positive weights and the dashed line represents negative weights.

\subsubsection{PCTL properties and results}
Then we checked two of the most important properties for contralateral inhibition to show that it is well implemented.   

\(P_{4}:\,P_{=?} [ F\,G\,(y_{2}=0 \wedge (F\, (y_{1} = 1))) ]\)
This specification ensures that in all future times, Neuron 2 stops spiking forever (becomes permanently inactive), and from that point onward, Neuron 1 continues to spike infinitely often, which means Neuron 1 is the winner and takes all priorities.

\(P_{5}:\, P_{>=1}[G\,>T\, (y_{2}=0)] \)
This specification ensures that in all possible evolutions of the system, after an initial period of T time steps where the system may oscillate, Neuron 2 ceases to spike entirely and remains inactive for all subsequent time. 

Table~\ref{tab:results-ci} shows the outcomes and execution time of \(P_4\) and \(P_5\).

\begin{table}[htbp]
\centering
\caption{Verification Results of Contralateral Inhibition}
\label{tab:results-ci}
\resizebox{\textwidth}{!}{%
\begin{tabular}{@{}cccc@{}}
\toprule
\textbf{Properties} & \textbf{Specifications} & \textbf{Outcomes} & \textbf{Time}\\
\midrule
\(P_{4}\)
& $P_{=?} [ F\,G\,(y_{2}=0 \wedge (F\, (y_{1} = 1))) ]$
& 1.0
& 372 ms\\
\midrule
\(P_{5}\)
& $ P_{>=1}[G\,>T\, (y_{2}=0)]$
& True (T=100)
& 165 ms\\
\bottomrule
\end{tabular}%
}
\end{table}

Those two properties demonstrate the correctness of our contralateral inhibition model and the fidelity of its biological simulation. These formal guarantees align with empirical observations of contralateral inhibitory circuits in sensory systems, where one neuronal population reliably dominates after mutual competition, thereby validating both the structural integrity and functional realism of our RP-L\&IF-based network model.  

\subsubsection{Nengo Simulation}
\begin{figure}[ht]
    \centering
    \begin{minipage}{0.45\textwidth}
        \centering
        \includegraphics[width=\linewidth]{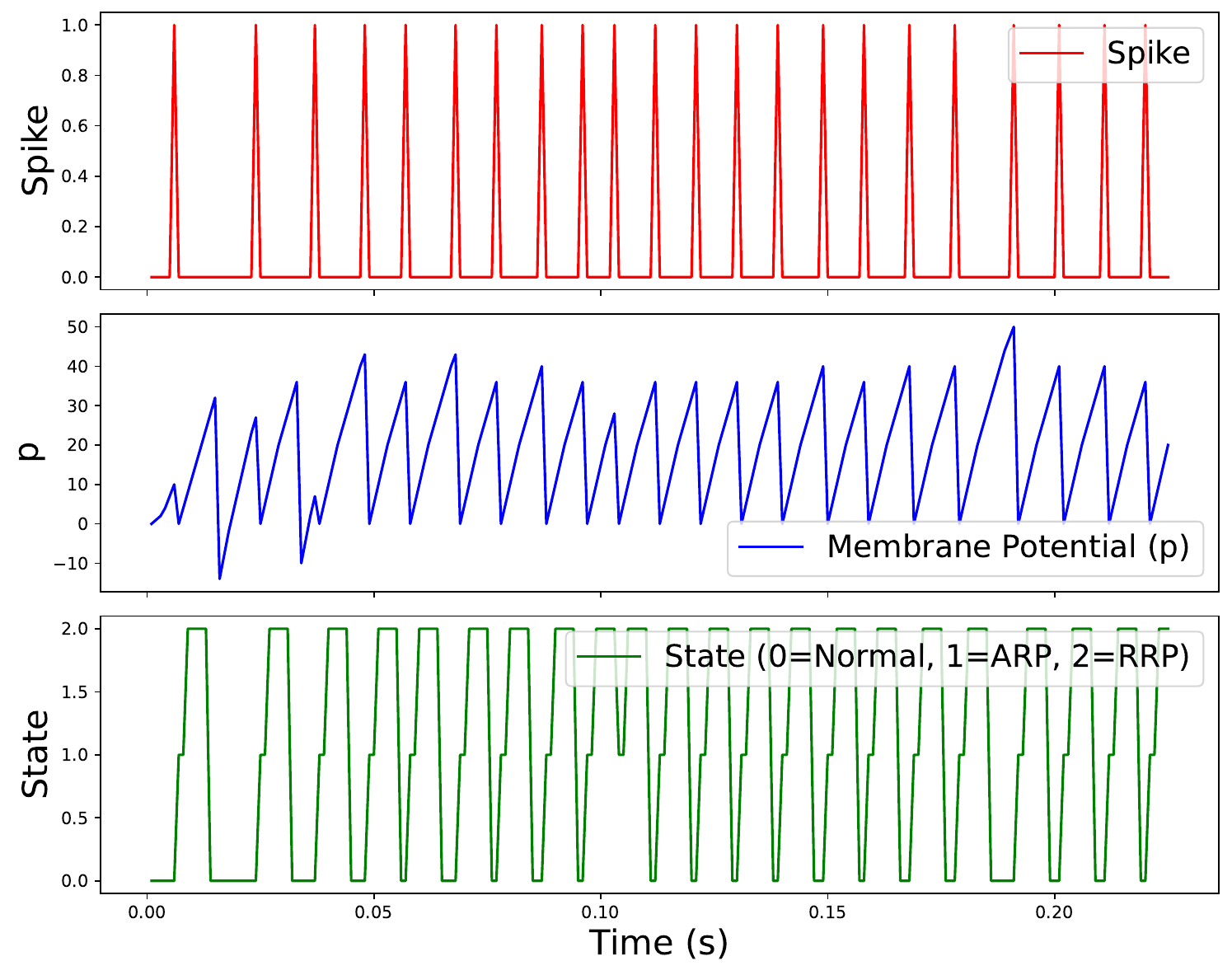}
        \subcaption{Data of Neuron1}
        \label{fig:ci-n1}
    \end{minipage}
    \hfill
    \begin{minipage}{0.45\textwidth}
        \centering
        \includegraphics[width=\linewidth]{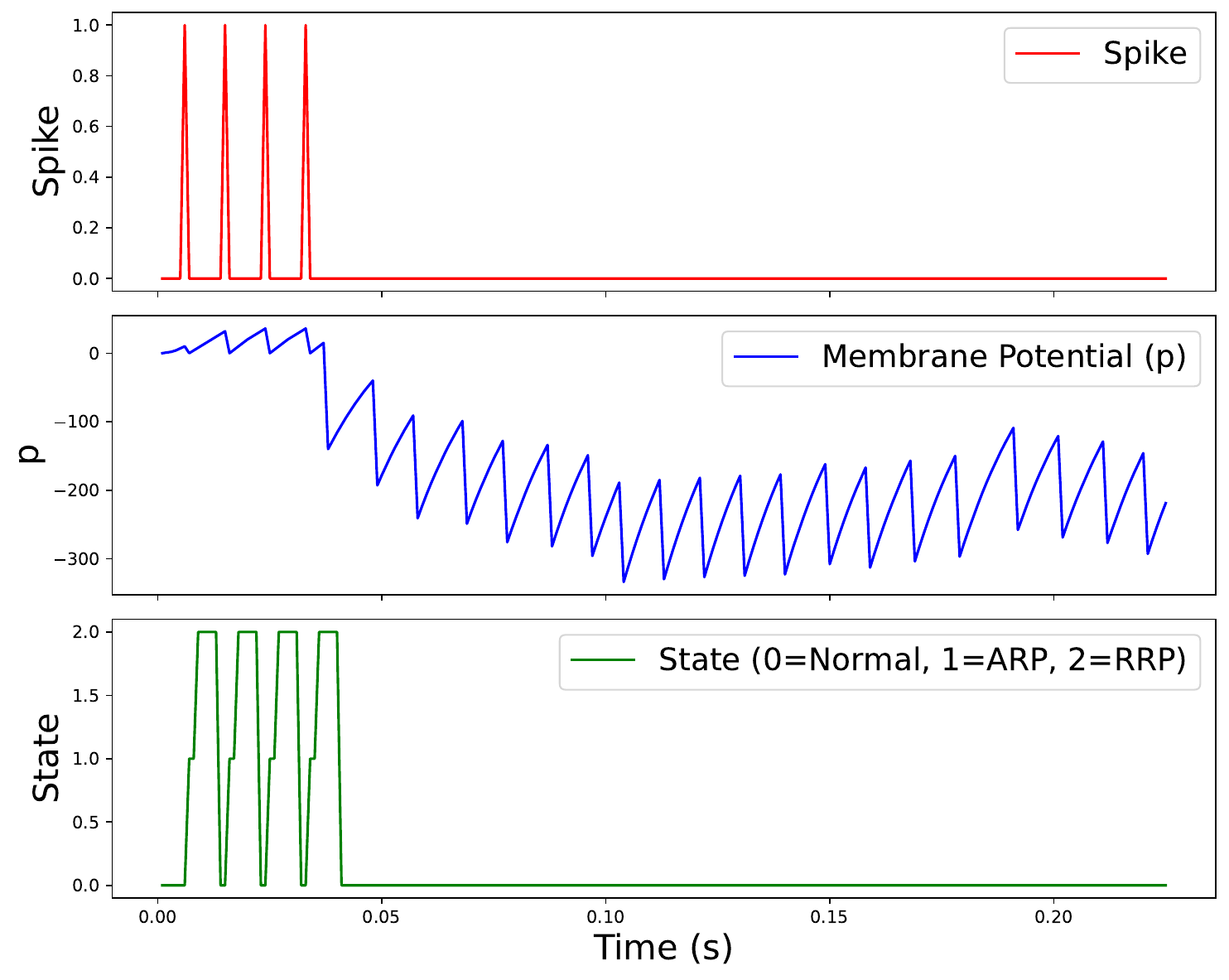}
        \subcaption{Data of Neuron2}
        \label{fig:ci-n2}
    \end{minipage}
    \caption{Simulation of Contralateral Inhibition }
    \label{fig:ci-data}
\end{figure}

Thanks to SNN-RF, we can automatically generate the Nengo RP-L\&IF neural network model from identical parameters and topology. For contralateral inhibition, our simulation results illustrate the evolution of each neuron's internal state variables alongside the PRISM model-checking outcomes rendered visually (see Figure~\ref{fig:ci-data}).

As shown in Figure~\ref{fig:ci-n2}, Neuron2 emits a single spike at the very start of the simulation and remains silent thereafter. In contrast, in Figure~\ref{fig:ci-n1}, Neuron1 continues to spike throughout the entire simulation, passing through its active period, normal period, and two refractory periods, with spikes occurring at irregular intervals.


\subsection{Convergent Excitatory Circuit}

\begin{figure}
    \centering
    \includegraphics[width=0.75\linewidth]{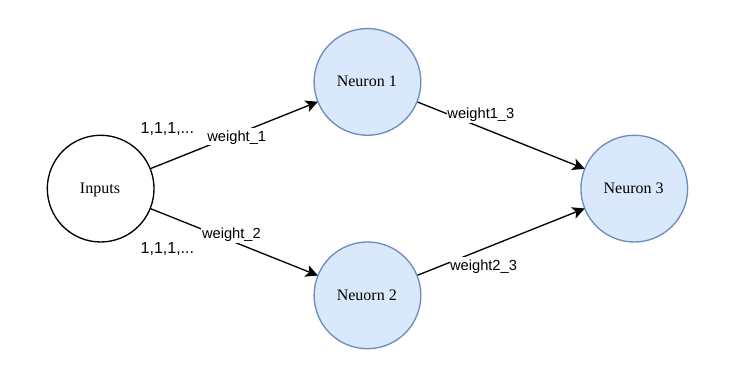}
    \caption{Convergent Excitatory Circuit}
    \label{fig:cec}
\end{figure}

Convergent excitatory circuits integrate multiple inputs to enhance downstream spiking, a mechanism crucial for rhythmic behaviors like locomotion~\cite{Grillner2006,Kiehn2016}. We model such a network with one constant input and two RP-L\&IF neurons exciting a postsynaptic target (Fig.~\ref{fig:cec}, solid lines indicate positive weights), to analyze temporal coherence using our framework.

\subsubsection{PCTL properties and results}
Here we verify three gating and activation properties under different synaptic configurations. Those two properties illustrate different types of convergent excitatory motifs:

\(P_{6}:\,P_{\ge 1}\bigl[(\,G(y_{2}=0)\wedge F(y_{1}=1)\,)\;\longrightarrow\;F(y_{3}=1)\bigr]\) asserts that with probability one, if Neuron 2 remains silent throughout and Neuron 1 spikes at least once, then Neuron 3 will eventually spike. It guarantees the conditional activation of Neuron 3 by Neuron 1 in the absence of activating input from Neuron 2.  

\(P_{7}:\,P_{\ge 1}\bigl[\,G\bigl((y_{1}=0\lor y_{2}=0)\bigr)\;\longrightarrow\;\neg F(y_{3}=1)\bigr]\) ensures that with probability one, whenever either Neuron 1 or Neuron 2 never spikes, Neuron 3 will never spike. It enforces that a single missing presynaptic activation is sufficient to block downstream firing.  

Table~\ref{tab:results-cec} shows the results and execution time of \(P_6\) and \(P_7\).

\begin{table}[htbp]
\centering
\caption{Verification Results of Convergent Excitatory Circuit}
\label{tab:results-cec}
\resizebox{\textwidth}{!}{%
\begin{tabular}{@{}cccc@{}}
\toprule
\textbf{Properties} & \textbf{Specifications} & \textbf{Outcomes} & \textbf{Time}\\
\midrule
\(P_{6}\)
& $P_{\ge 1}\bigl[(\,G(y_{2}=0)\wedge F(y_{1}=1)\,)\;\longrightarrow\;F(y_{3}=1)\bigr]$
& True
& 9 ms\\
\midrule
\(P_{7}\)
& $P_{\ge 1}\bigl[\,G\bigl((y_{1}=0\lor y_{2}=0)\bigr)\;\longrightarrow\;\neg F(y_{3}=1)\bigr]$
& True 
& 1 ms\\
\bottomrule
\end{tabular}%
}
\end{table}

\subsubsection{Nengo Simulation}
\begin{figure}
    \centering
    \begin{minipage}{0.45\textwidth}
        \centering
        \includegraphics[width=\linewidth]{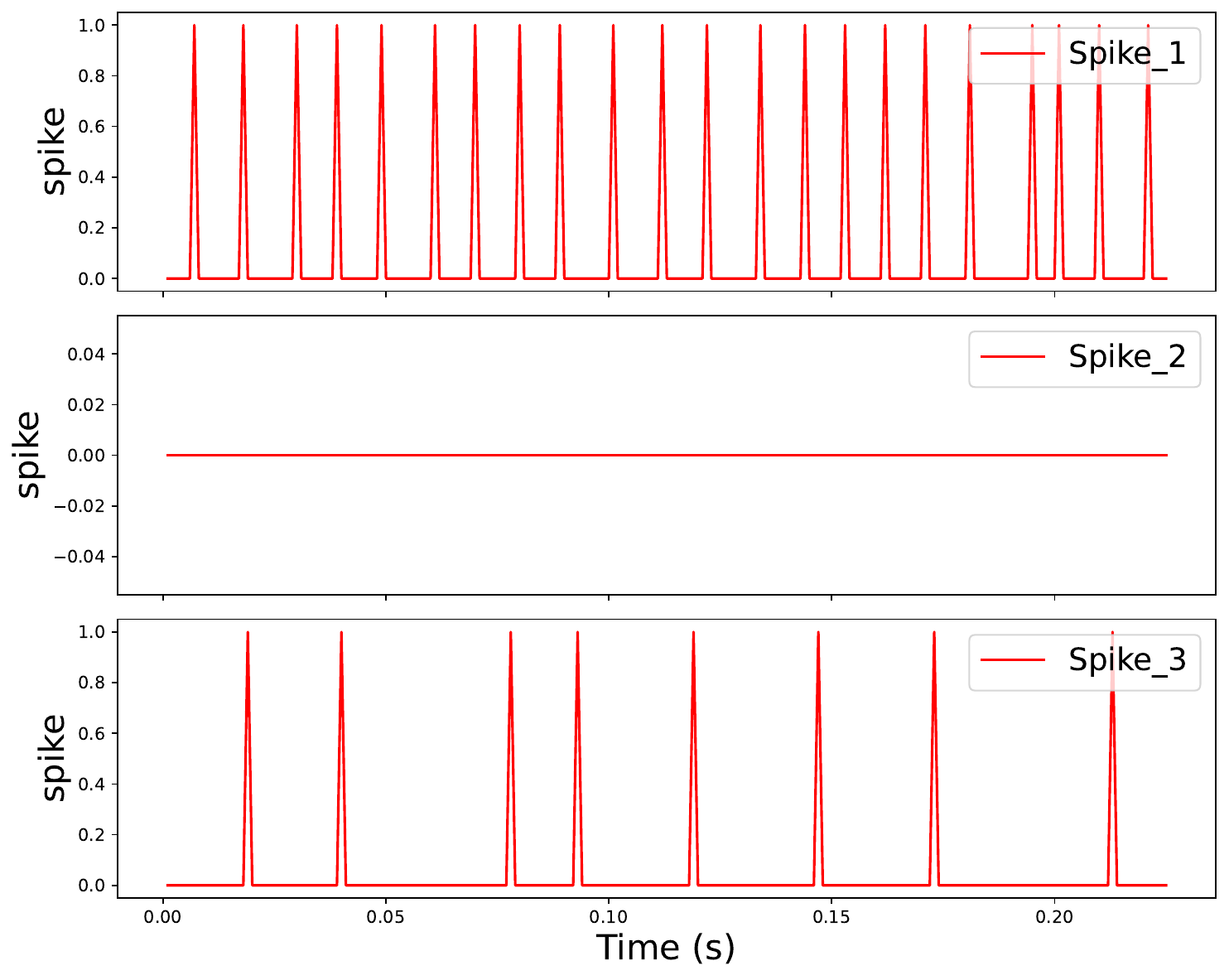}
        \subcaption{Simulation of neuron 1, 2, and 3 with parameters respecting \(P_6\)}
        \label{fig:cec-p1}
    \end{minipage}
    \hfill
    \begin{minipage}{0.45\textwidth}
        \centering
        \includegraphics[width=\linewidth]{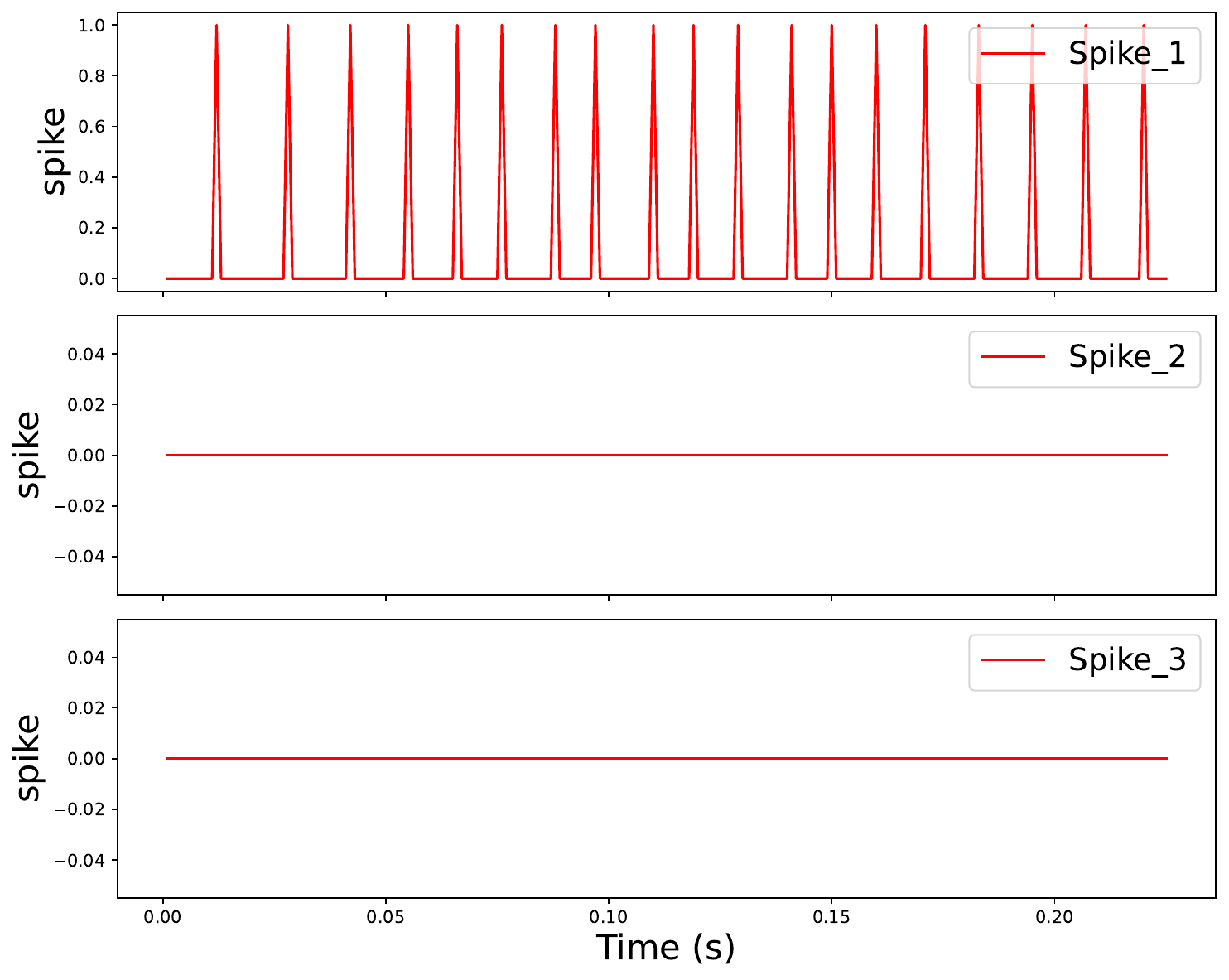}
        \subcaption{Simulation of neuron 1, 2, and 3 with parameters respecting \(P_7\)}
        \label{fig:cec-p2}
    \end{minipage}
    \caption{Simulation of Convergent Excitatory Circuit (the X-axis represents time, and the Y-axis represents spike}
    \label{fig:cec-sim}
\end{figure}

In the Nengo part, we simulate two synaptic configurations that satisfy the desired properties, respectively. To better demonstrate our simulation results, here we display the comparisons of spikes of each neuron (see Figure~\ref{fig:cec-sim}).

Property 6 ensures Neuron 3 activates with just one presynaptic spike, mimicking the brain's reflex-like response to stimuli (Figure~\ref{fig:cec-p1}). Property 7 requires spikes from both Neuron 1 and Neuron 2 for Neuron 3 to spike, akin to decision-making needing dual sensory inputs (Figure~\ref{fig:cec-p2}).


\section{Conclusion and Future Work}\label{sec:conclusion}

SNNs ambition to represent natural neuromorphic computations. Precise representation of basic neuron archetypes' spatiotemporal dynamic (spike timing/probability), as well as of means to combine them into larger brain-like systems, should allow checking how (elementary) properties of the former can contractually ensure more global (stochastic) requirements on the latter, using formal verification (PRISM) for medium-scale proofs and stochastic simulation (Nengo) for large-scale validation.

This paper focuses on the probabilistic temporal specification of individual neuron archetypes. We introduced the RP-LI\&F neuron model, unifying discrete-time refractory periods with probabilistic spike generation for biological plausibility. The SNN-RF specification enables automated co-verification across PRISM and Nengo, eliminating toolchain inconsistencies. Case studies on biological archetypes like contralateral inhibition demonstrate preserve neural dynamics with probabilistic formal assurances, validating our co-design approach.

As for future work, the main kind of properties we will want to check are the following contractual requirements: assuming that elementary neuron archetypes are sufficiently alert (in timing and probability of reactive spiking), then the global brain-like neuromorphic computation is guaranteed to satisfy similar reactive constraints. But we want to rely on existing software frameworks to perform these property analyses, namely PRISM and Nengo, and they do not offer such an assume/guarantee contract approach natively. Dealing with the assumption part is rather simple, as it just means to instantiate the parameters to their desired values; note that the SNN-RF format will allow for easy change of such parameter values without affecting the rest of the design specification. Dealing with global requirement constraints is much more involved and largely a matter of future work. Currently, we still write PCTL formulae directly in and for PRISM. The case for Nengo would be to add side-observers running alongside the simulation, and recording spiking test results; the issue then is to provide the number of simulation runs needed (in a Monte-Carlo fashion) to obtain statistically reasonable confidence for our constraint satisfaction.

Finally, based on ~\cite{DeMaria2020}, we intend to propose an algorithm to find parameters such that some key dynamic properties of SNNs are verified in a probabilistic context. 
\bibliographystyle{splncs04}
\bibliography{mybib}
%

\end{document}